%% file: main.tex
\ifdefined\arxivversion
\else
\def\cvprversion{1}
\fi
\ifdefined\cvprversion
\documentclass[10pt,twocolumn,letterpaper]{article}
\usepackage[T1]{fontenc}
\usepackage[pagenumbers]{cvpr}
\else
\documentclass[11pt,letterpaper]{article}
\usepackage[T1]{fontenc}
\usepackage[margin=1in]{geometry}
\usepackage{lmodern}
\usepackage{authblk}
\usepackage{graphicx}
\usepackage{amsmath,amssymb}
\usepackage[numbers,sort&compress]{natbib}
\usepackage[hyphens]{url}
\usepackage[font=small,labelfont=bf]{caption}
\usepackage{subcaption}
\usepackage[shortlabels,inline]{enumitem}
\def\arxivversion{1}
\fi
\usepackage{microtype}
\ifdefined\revisionclean\else
\def\revisioninline{1}
\fi
\def\revisionclean{1}
\input{preamble}

\usepackage[breaklinks,colorlinks,linkcolor=black,citecolor=black,urlcolor=blue]{hyperref}
\usepackage[capitalize]{cleveref}
\hypersetup{pdftitle={How Far Can Coding Agents Go in Visual Generation?},pdfauthor={Hengguang Zhou, Yuanhao Ban, Jiaqi Feng, Xiaohuan Pei, Justin Cui, and Cho-Jui Hsieh}}
\title{How Far Can Coding Agents Go in Visual Generation?}
\ifdefined\cvprversion
\author{%
Hengguang Zhou\textsuperscript{1}\thanks{These authors contributed equally to this work.}\quad
Yuanhao Ban\textsuperscript{1*}\quad
Jiaqi Feng\textsuperscript{2}\\
Xiaohuan Pei\textsuperscript{1}\quad
Justin Cui\textsuperscript{1}\quad
Cho-Jui Hsieh\textsuperscript{1}\\[0.4em]
\textsuperscript{1}University of California, Los Angeles\qquad
\textsuperscript{2}Tsinghua University
}
\else
\author[1]{Hengguang Zhou\thanks{These authors contributed equally to this work.}}
\author[1]{Yuanhao Ban\textsuperscript{\ensuremath{*}}}
\author[2]{Jiaqi Feng}
\author[1]{Xiaohuan Pei}
\author[1]{Justin Cui}
\author[1]{Cho-Jui Hsieh}

\affil[1]{University of California, Los Angeles}
\affil[2]{Tsinghua University}
\fi

\date{}

\makeatletter
\providecommand{\@LN@col}[1]{}
\providecommand{\@LN}[2]{}
\ifdefined\cvprversion
\g@addto@macro\@maketitle{\input{sec/0_teaser}}
\fi
\makeatother

\begin{document}
\maketitle
\ifdefined\cvprversion\else
\input{sec/0_teaser}
\fi
\input{sec/0_abstract}
\input{sec/1_intro}
\input{sec/2_related}
\input{sec/3_settings}
\input{sec/4_zeroshot}
\input{sec/5_harness}
\input{sec/7_internal_visual_thought}
\input{sec/7_discussion}
\FloatBarrier
{
    \small
    \ifdefined\cvprversion
    \bibliographystyle{ieeenat_fullname}
    \else
    \bibliographystyle{unsrtnat}
    \fi
    \bibliography{main}
}
\clearpage
\input{sec/appendix.tex}

\end{document}

%% file: preamble.tex
\usepackage{booktabs}
\usepackage{xcolor}
\input{revision_markup}

\usepackage{dblfloatfix}
\usepackage{placeins}

%% file: revision_markup.tex
\newif\ifrevisionmarkup
\revisionmarkuptrue
\ifdefined\revisionclean\revisionmarkupfalse\fi
\definecolor{revisionbefore}{HTML}{A12C36}
\definecolor{revisionafter}{HTML}{174EA6}
\definecolor{revisionnote}{HTML}{50565D}

\DeclareRobustCommand{\revblue}[1]{%
  \ifdefined\revisioninline
    {\color{revisionafter}#1}%
  \else
    #1%
  \fi
}

\newcommand{\revadd}[1]{%
  \ifrevisionmarkup
    \par\begingroup\color{revisionafter}%
    \noindent\textbf{[Added]} #1\par\endgroup
  \else
    #1%
  \fi
}

%% file: sec/0_teaser.tex
\noindent\begin{minipage}{\textwidth}
\centering
\includegraphics[width=\linewidth]{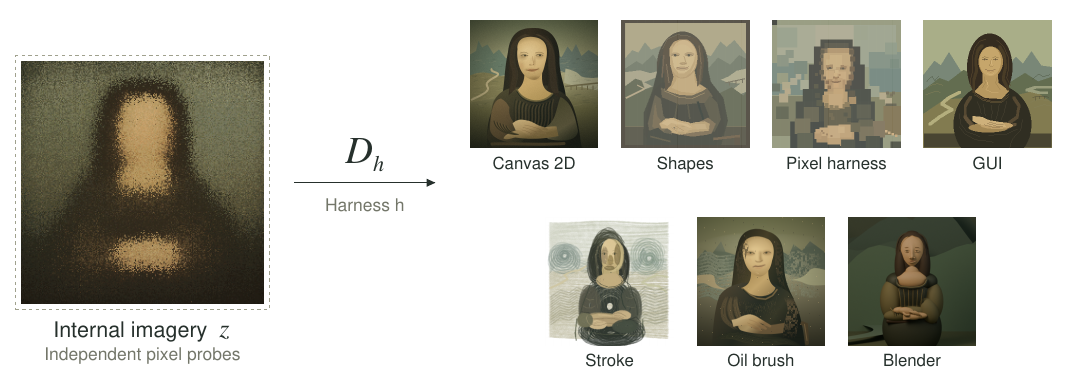}
\captionof{figure}{A conceptual view of shared internal imagery across drawing harnesses. Left: the Mona Lisa assembled from 65,536 independent fresh-session pixel queries on a $256\times256$ grid, used as a behavioral probe of hypothesized internal imagery $z$. Right: one-shot $256\times256$ whole-image outputs through seven drawing harnesses, generated without visual feedback. The mapping $I_h = D_h(z)$ illustrates the hypothesis; the arrow denotes a conceptual relation rather than an experimental image-to-image pipeline.}
\label{fig:teaser}
\end{minipage}
\par\vspace{1em}

%% file: sec/0_abstract.tex
\begin{abstract}
Modern image generation largely relies on specialized generative models that leave the process of image construction implicit in learned parameters. We explore an alternative: general-purpose coding agents that construct images through executable programs. Our study of zero-shot drawing on AgentDrawBench, a 200-prompt benchmark, reveals a distinctive capability profile. Agents excel at preserving structure and satisfying precise visual constraints, achieving competitive results in domains where compositional accuracy matters more than photorealistic appearance, while remaining limited in perceptual richness. Their visual expressiveness varies with the drawing interface, and a qualitative study on reconstruction shows how procedural skills extracted from human demonstrations support different artistic media. We then examine code as an external visual language of thought: a structured, executable representation that agents can construct and inspect. We test it in visual question answering task and find reconstructed visual code retains substantial information for question answering and, in some domains, supports higher accuracy than the original image. These abilities raise a further question: do agents draw on a shared internal representation of scene content beyond a particular drawing program? We explore an internal visual language-of-thought hypothesis, in which a shared, compositionally structured representation supports visual operations across drawing languages. Cross-harness drawings and images assembled from independent single-pixel queries exhibit related layouts and tones. Agents also maintain self-generated room arrangements across viewpoints after visible drawings are removed. These results demonstrate that the coding agent would maintain a consistent and coherent world inside its imagine even before the harness gets involved and the generation starts.
\end{abstract}

%% file: sec/1_intro.tex
\section{Introduction}
\label{sec:intro}
In recent years, text-to-image generation has largely relied on end-to-end generative models, including latent diffusion models, rectified-flow transformers, and autoregressive models~\cite{rombach2022latentdiffusion,saharia2022imagen,esser2024rectifiedflow,yu2022parti}.
Despite their considerable success, these approaches encode the generation process in learned neural networks, leaving the details of image construction largely implicit in model parameters. In contrast, modern agents offer a programmatic alternative: they can write and execute programs, coordinate tools, and iteratively refine their actions over multiple turns~\cite{liang2022codepolicies,yao2022react}.
These developments motivate our central question:
\textbf{How far can general-purpose coding agents go in visual generation?}

We first assess these agents as zero-shot drawers using harnesses ranging from individual pixels and geometric primitives to Canvas~2D programs, SVG documents, and a graphical painting application. To evaluate these coding agents, we introduce AgentDrawBench, a 200-prompt benchmark spanning multiple domains. We will publicly release the benchmark to support further research on agent-based visual generation. Our results show that these agents are strong in structural fidelity but fall short in perceptual realism compared with native image generators. In particular, GPT-6-Astra with Canvas~2D achieves 85.8\% structural fidelity, compared with 68.8\% for \texttt{gpt-image-1} and 90.7\% for \texttt{gpt-image-2}. On design and symbolic tasks, where precise control matters more than photorealistic appearance, GPT-6-Astra with Canvas~2D achieves a 92.0\% holistic win rate against \texttt{gpt-image-1}, judged by \texttt{gemini-3.5-flash}. This task-dependent profile reveals the potential of coding agents for visual generation without a dedicated image generator. These results raise a further question: can agents also use visual programs to organize and reason about the scenes they depict?

The language-of-thought hypothesis proposes that thinking operates over structured mental representations whose meanings depend on their components and how they are combined~\cite{rescorla2024language}. Drawing on this perspective, we frame visual code as an \emph{external visual language of thought}: a structured, executable medium for organizing, manipulating, and reasoning about visual content. We examine whether code can serve this role in both visual generation and understanding. For generation, the agent achieves 83.0\% structural fidelity using code alone, close to the 85.6\% achieved when it can inspect intermediate images. For understanding, we sample 200 visual question answering tasks from WorldBench~\cite{yin2026worldbench} covering a broad taxonomy of real-world visual concepts and reconstruct the question images as SVG code.
We find that answering from source code achieves 69.0\% accuracy, compared with 65.0\% from the render and 74.0\% from the original image. In the academics and digital-world domains, the agent answers questions more accurately using code than using the original images.


With code serving as an external visual language of thought, we next ask whether capabilities that emerge in large language models translate to drawing with code. Specifically, we examine whether code-based drawing benefits from test-time scaling and in-context learning using procedural drawing guidance from human demonstrations. We find that extending refinement to a fourfold budget yields a modest structural improvement and only a 50.5\% holistic win rate against the standard run. Analysis of the outputs points to the importance of the drawing interface. For in-context learning, we provide procedures distilled from human drawing demonstrations, including stroke operations, construction order, value grouping, and edge control, enhancing the drawing interfaces. Guided by these procedures, the agent reconstructs reference images in Chinese ink, charcoal, watercolor, and oil painting, producing expressive, medium-specific visual effects. These experiments highlight the role of drawing interfaces in perceptual quality. They also illustrate how code can serve as a working representation for visual thinking, enabling agents to apply medium-specific drawing procedures supplied in context.

Across these external expressions, a recurring pattern emerges: independent drawings of the same prompt can share coarse layouts and color choices across different harnesses (\cref{fig:harness-ladder}). Could these recurring visual choices reflect a shared internal representation of scene content? We formulate this possibility as an \emph{internal visual language-of-thought hypothesis}: agents may use a compositionally structured representation of entities, attributes, and relations to construct, interpret, and transform scenes through different external expressions~\cite{rescorla2024language,quiltydunn2023best}. The hypothesis motivates two behavioral properties: \emph{consistency}, the recurrence of scene-level choices across forms of expression, and \emph{coherence}, the coordination of entities and relations under visual transformations.

We investigate these properties through complementary probes. We quantify cross-harness consistency and test independent single-pixel queries, each made without access to any other query's answer. The assembled images recover regions and tones similar to whole-image drawings. To examine spatial coherence, agents choose furniture arrangements and revisit them from different viewpoints. Scene organization remains largely consistent after earlier SVGs are removed, including a condition that also removes the furniture list. Fresh calls given only the furniture list and turn history show lower agreement across revisits and between views and floor plans. Together, the results provide complementary behavioral support for the internal visual language-of-thought hypothesis.

Our contributions are:
\begin{itemize}
  \item We characterize agents as zero-shot drawers, identifying strong structural fidelity and precise control alongside limits in perceptual realism, and examine how feedback, refinement, and drawing interfaces affect their performance (\cref{sec:main,sec:harness-effects}).
  \item We study code as an external visual language of thought, showing how reconstructed SVG programs support question answering about scene content (\cref{sec:representation}).
  \item We formulate an internal visual language-of-thought hypothesis and investigate complementary evidence for reusable scene content through cross-harness consistency, independent pixel queries, and spatial coherence in self-generated rooms (\cref{sec:inner-vlot}).
\end{itemize}

%% file: sec/2_related.tex
\section{Related Work}
\label{sec:related}

\paragraph{Text-to-image generative models.}
Diffusion models have achieved substantial success in text-to-image synthesis in recent years~\cite{stability2024sd35,bfl2026flux2,openai2025gptimage,google2025nanobanana,bytedance2025seedream4}. \cite{rombach2022latentdiffusion} propose to diffuse the latent instead of the image pixels~\cite{rombach2022latentdiffusion}.
Imagen combines language
representations with cascaded diffusion~\cite{saharia2022imagen}. \citet{esser2024rectifiedflow} transform random noise into data along nearly straight paths, making generation simpler and faster. \citet{ho2022classifier} use the difference between conditional and unconditional noise predictions to guide generation toward alignment with the input prompt. Transfusion~\citet{zhou2025transfusion} unified image tokens and language tokens in one shared transformer and provides better control.

\paragraph{Code generation for visual content.}
Graphics programs represent images through executable
descriptions~\cite{ellis2018learning}.
StarVector~\cite{rodriguez2025starvector},
OmniSVG~\cite{yang2026omnisvg}, and
InternSVG~\cite{wang2026internsvg} learn models for structured
vector generation.
Stroke-based reconstruction is also studied through learned
painting policies with neural renderers~\cite{huang2019learning}
and feed-forward prediction of stroke sets~\cite{liu2021painttransformer}.
SketchAgent uses an off-the-shelf multimodal language model and
an in-context sketching language to construct and revise drawings
stroke by stroke~\cite{vinker2025sketchagent}.

\paragraph{Agentic image-generation systems.}
Agents organize image creation through tool selection, external
knowledge gathering, and iterative refinement.
Early systems coordinate visual foundation models~\cite{wu2023visualchatgpt}, select suitable
diffusion generators~\cite{qin2024diffusiongpt},
and revise generation prompts using visual feedback~\cite{yang2023idea2img}.
Recent work extends these capabilities. VisionDirector decomposes
long instructions into structured goals and verifies staged edits
with rollback~\cite{chu2025visiondirector}.
For knowledge-intensive requests, Gen-Searcher performs 
search to gather textual knowledge and reference images~\cite{feng2026gensearcher}.
RS-Gen uses a training-free questioning-and-solving loop for
reasoning and search-augmented generation~\cite{bian2026rsgen}. As a related concurrent work, MaLiang-Harness studies programmable image and video generation, addressing the gap between executable programs and their visual outcomes~\cite{zhao2026maliangharness}.

%% file: sec/3_settings.tex
\section{Experimental Setup}

\label{sec:setup}
In this section, we describe various drawing harnesses, AgentDrawBench, and the structural, perceptual, and holistic metrics.

\subsection{Harness Design}
\label{sec:agent} 
We design several harnesses for programmable drawing, ranging from direct pixel assignments and shape-based marks to executable Canvas 2D programs and SVG documents.  To evaluate whether agents can draw in a human-like manner, we also assess their ability to imitate human drawing strokes and draw through graphical user interfaces (GUIs).

\begin{itemize}
  \item \textbf{pixel:} The model directly assigns the location and color of each pixel on a persistent canvas autoregressively. Previously painted pixels can be overwritten in later turns. No geometric primitives or procedural drawing operations are available. This provides the least abstract output
  representation, but its generation cost grows with the number of pixels.

  \item \textbf{shapes:} The model constructs an image as a
  sequence of elementary geometric marks: filled rectangles,
  ellipses, polygons, lines, and individual pixels. It predicts each mark's position, extent, color, and stroke width. Marks are rasterized in generation order,
  allowing later shapes to cover earlier ones and progressively build a composition from background to foreground.

  \item \textbf{canvas2d:} The model generates executable
  JavaScript programs that draw onto a persistent canvas.
  Each program can combine paths, curves, text, gradients,
  transformations, and compositing operations, while loops and
  helper routines express repeated objects or
  procedural textures. A persistent state object can retain reusable
  drawing routines and intermediate data across turns, enabling complex, high-resolution compositions without enumerating every drawing operation separately.

  \item \textbf{svg:} The model represents the entire image as
  a structured vector document, specifying shapes, paths, text,
  gradients, transformations, and their ordering declaratively.
  After that, a renderer converts this document into the output image.

  \item GUI:\label{sec:gui-protocol}
  The model operates JS Paint, a graphical painting application,
  through mouse clicks, drags, keyboard input, and color selection.
  It selects drawing tools in the interface and constructs the
  image on the application's persistent canvas. After each action
  batch, a screenshot of the full window provides feedback for
  correcting tool selection, geometry, and placement. The agent
  executes no drawing code and cannot paste or import images.
  Interface and rollout settings are in \cref{sec:app-gui}.

  \item Blender Cycles:\label{sec:blender-protocol}
  The model constructs a three-dimensional scene through Python,
  specifying geometry, materials, lights, and a camera. Cycles
  computes projection, shading, and light transport~\cite{blender2026cycles}.
  The agent inspects preview renders and revises its scene across
  turns; each script rebuilds the scene in a fresh process.
  External models, images, and material libraries are excluded.
  The model therefore specifies the scene assets while the renderer
  computes their appearance. Detailed settings are in
  \cref{sec:app-blender-settings}.

  \item Stroke workflows: Medium-specific drawing procedures
  guide reference-based reconstruction on a persistent canvas.
  The protocol and four variants are described in
  \cref{sec:style-transfer}.

\end{itemize}

\begin{figure*}[!t]
\centering
\includegraphics[width=\textwidth,height=0.55\textheight,keepaspectratio]{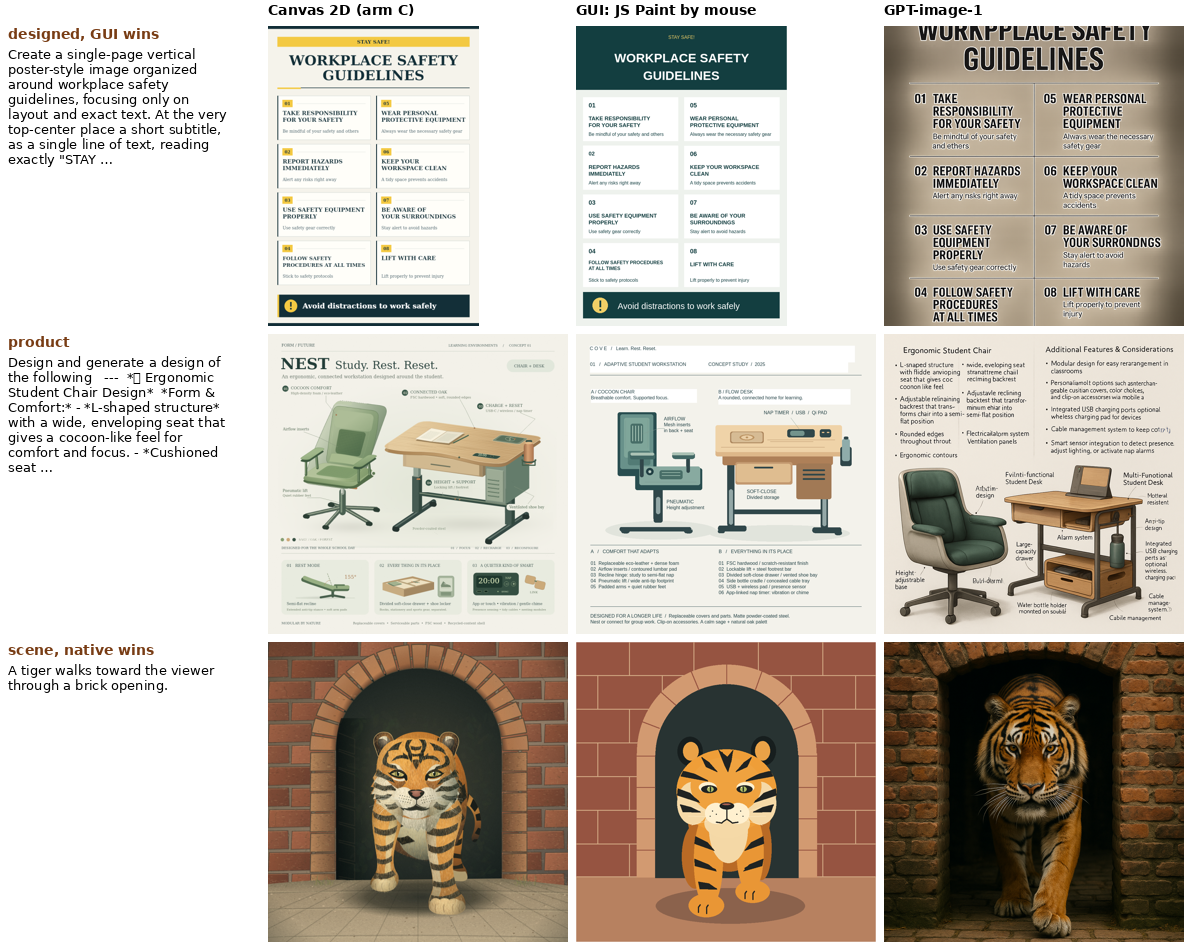}
\caption{The computer-use harness. Three matched AgentDrawBench prompts drawn through Canvas~2D with visual feedback, the JS Paint GUI, and GPT-image-1: workplace-safety design, student furniture, and a tiger approaching through a brick opening.}
\label{fig:gui-harness}
\end{figure*}

\Cref{fig:gui-harness} compares Canvas~2D and GUI drawings against native image generation on the same prompts. The primary generator is \texttt{gpt-6-astra}. Prompts, tool execution, stopping rules, and budget controls are detailed in \cref{sec:app-harness,sec:app-skeleton}.

\subsection{AgentDrawBench}
\label{sec:benchmark}
\paragraph{Benchmark Construction} We construct AgentDrawBench, a 200-prompt benchmark across seven families: photographic scene, human/organic,
product/material, spatial/3D, stylized illustration, designed visual,
diagram/symbolic. We select prompts from four published benchmarks: 110 from Arena-T2I-Hard~\cite{ban2026arena}, 40 from HPDv3~\cite{ma2025hpsv3}, 35 from BizGenEval~\cite{li2026bizgeneval} and 15 from GraphicWeaver~\cite{ki2026graphicweaver}. We evaluate structural fidelity, perceptual realization, and holistic quality on AgentDrawBench. We will publicly release AgentDrawBench.

\paragraph{Structural fidelity.} We assess prompt adherence using question checklists about the
generated image. We retain the source benchmarks' question sets
where available: DSG-style questions~\cite{cho2024davidsonian} from Arena-T2I-Hard~\cite{ban2026arena},
20-question checklists from BizGenEval~\cite{li2026bizgeneval}, and binary VQA questions
from GraphicWeaver~\cite{ki2026graphicweaver}. For HPDv3~\cite{ma2025hpsv3}, we generate DSG-style atomic
questions from each prompt and manually audit the resulting
checklists. These questions cover requested objects, attributes,
counts, spatial relationships, text, and layout. For each image,
a VLM answers every question with \emph{yes}, \emph{no}, or
\emph{unclear}, accompanied by visual evidence. We score questions
independently and report the proportion answered \emph{yes},
treating both \emph{no} and \emph{unclear} as failures.
For Arena, the structural comparison retains only the source
questions labeled faithfulness.

\paragraph{Perceptual quality.}
An image can satisfy the requested content yet appear visually unconvincing. We therefore assess perceptual quality through relative comparisons on the same prompt, asking which aspects of an agent's output fall short of a native image-generation baseline. The perceptual rubric contains ten dimensions: artifacts, object form, anatomy, typography, perspective,
light transport, material, scene physics, medium realization and semantic coherence. For each dimension applicable to a prompt, a VLM compares the
two outputs and selects the better image. To eliminate position bias, each comparison is
performed twice with image positions swapped; a win requires
agreement across both orderings, while disagreements are
reported as ties.

\paragraph{Holistic quality.} We adopt the MMRB2 pairwise judge prompt~\cite{hu2026multimodal} to assess holistic quality. Each pair is judged twice with positions swapped.
We report preference win rates as $(W+0.5T)/(W+T+L)$, assigning half credit to ties and disagreements between the two image orderings. A win requires agreement across both image orderings; failed or incomplete judgments are excluded and their coverage is disclosed.

%% file: sec/4_zeroshot.tex
\section{Coding Agents Are Zero-Shot Drawers}
\label{sec:main}

We first evaluate coding agents as zero-shot drawers: how faithfully can they depict requested content, how convincingly can they realize it, and how often are their outputs preferred to a native image generator? Agents receive tool documentation and a text prompt, without task-specific fine-tuning or example drawings; multiple turns and feedback from their own renders remain available. We report structural fidelity, perceptual realization, and holistic preference across agents, task categories, and drawing harnesses. These comparisons characterize zero-shot drawing capability before we examine how agents construct and understand visual content through code in \cref{sec:representation}.

\paragraph{Structural Fidelity}
\label{sec:structural-control}
\Cref{tab:structural} reports checklist pass rates 
on AgentDrawBench, both overall and across content categories. Agent systems achieve strong structural fidelity: GPT-6-Astra attains an overall pass rate of 85.8\%, outperforming \emph{gpt-image-1} by 17 percentage points. Performance is particularly strong in the design and symbolic categories, where zero-shot DeepSeek V4.1 Flash achieves a pass rate of 89.5\%, exceeding \emph{gpt-image-1} by 31.4 percentage points. In the design and symbolic categories, Astra also slightly exceeds the structural pass rate of \emph{gpt-image-2}, the higher-scoring native baseline overall.

\begin{table}[t]
\centering
\small
\setlength{\tabcolsep}{1.5pt}
\resizebox{\columnwidth}{!}{%
\begin{tabular}{lcccccccc}
\toprule
Harness & \multicolumn{3}{c}{Canvas 2D} & Shapes & SVG & Blender & \multicolumn{2}{c}{Native} \\
\cmidrule(lr){2-4}\cmidrule(lr){5-5}\cmidrule(lr){6-6}\cmidrule(lr){7-7}\cmidrule(lr){8-9}
Agent & Astra & Kimi K3 & \shortstack{DeepSeek} & Astra & Astra & Astra & \shortstack{GPT-image-2} & \shortstack{GPT-image-1} \\
\midrule
Scenes & 76.5 & 52.8 & 58.8 & 68.1 & 71.9 & 62.0 & 85.4 & 76.8 \\
People / organic & 81.1 & 54.7 & 69.4 & 71.4 & 79.1 & 73.2 & 93.6 & 80.1 \\
Objects / 3D & 82.0 & 60.5 & 70.3 & 63.3 & 77.8 & 71.5 & 87.0 & 67.1 \\
Stylized illustration & 88.5 & 49.3 & 73.6 & 82.3 & 85.8 & 71.9 & 92.4 & 74.7 \\
Design / symbolic & 94.6 & 84.7 & 89.5 & 75.2 & 93.9 & 92.3 & 94.4 & 58.1 \\
\midrule
Overall & 85.8 & 64.6 & 74.7 & 71.5 & 83.2 & 76.8 & 90.7 & 68.8 \\
\bottomrule
\end{tabular}}
\caption{Structural fidelity (\%) by pass rate. Astra denotes GPT-6-Astra; Deepseek denotes Deepseek V4.1 Flash. }
\label{tab:structural}
\end{table}

\paragraph{Perceptual Realization}
\label{sec:perceptual-limits}
\Cref{tab:capmap} reports perceptual win rates against GPT-image-1 across ten dimensions, judged by \texttt{gemini-3.5-flash}. Strong prompt adherence does not guarantee convincing appearance. With Astra, Canvas~2D and SVG achieve typography win rates of 77.6\% and 74.8\%, respectively, but all reported configurations remain below 50\% on anatomy, perspective, light transport, material, and scene physics. The agents can produce legible design elements while struggling with naturalistic form and appearance.

\begin{table}[t]
\centering
\small
\setlength{\tabcolsep}{3pt}
\resizebox{\columnwidth}{!}{%
\begin{tabular}{lcccccc}
\toprule
Harness & \multicolumn{3}{c}{Canvas 2D} & Shapes & SVG & Blender \\
\cmidrule(lr){2-4}\cmidrule(lr){5-5}\cmidrule(lr){6-6}\cmidrule(lr){7-7}
Agent & Astra & Kimi K3 & \shortstack{DeepSeek} & Astra & Astra & Astra \\
\midrule
artifacts & 45.2 & 32.3 & 29.0 & 28.7 & 48.9 & 39.7 \\
object form & 46.1 & 24.3 & 25.3 & 30.2 & 39.0 & 32.6 \\
anatomy & 15.4 & 9.9 & 8.2 & 10.4 & 18.9 & 13.6 \\
typography & 77.6 & 57.6 & 61.2 & 38.1 & 74.8 & 58.6 \\
perspective & 28.0 & 14.8 & 15.1 & 11.6 & 23.2 & 25.8 \\
light transport & 18.0 & 13.1 & 15.3 & 6.2 & 19.6 & 24.0 \\
material & 13.1 & 7.9 & 10.5 & 3.5 & 12.6 & 15.2 \\
scene physics & 28.0 & 14.1 & 13.8 & 17.0 & 23.2 & 24.2 \\
medium realization & 35.3 & 29.0 & 28.0 & 21.0 & 38.0 & 29.8 \\
semantic coherence & 51.5 & 28.8 & 31.1 & 35.6 & 45.2 & 39.4 \\
\bottomrule
\end{tabular}}
\caption{Perceptual win rates (\%) against GPT-image-1. Astra denotes GPT-6-Astra; Deepseek denotes Deepseek V4.1 Flash.}
\label{tab:capmap}
\end{table}

\paragraph{Holistic Performance}
\label{sec:holistic-performance}
We compare each agent configuration directly against GPT-image-1 using the MMRBv2 rubric with Gemini~3.5~Flash. \Cref{tab:holistic} shows the win rate over GPT-image-1 across different categories. Astra with Canvas~2D achieves a 92.0\% win rate on design and symbolic prompts, but only 13.0\% on scenes and 6.0\% on people and organic subjects. Its overall win rate is 47.7\%, counting ties as half a win. These results establish a pronounced task profile: agents are competitive on briefs emphasizing explicit structure and legible design, while the native baseline is preferred for naturalistic scenes and people.

\begin{table}[t]
\centering
\small
\setlength{\tabcolsep}{3pt}
\resizebox{\columnwidth}{!}{%
\begin{tabular}{lcccccc}
\toprule
Harness & \multicolumn{3}{c}{Canvas 2D} & Shapes & SVG & Blender \\
\cmidrule(lr){2-4}\cmidrule(lr){5-5}\cmidrule(lr){6-6}\cmidrule(lr){7-7}
Agent & Astra & Kimi K3 & \shortstack{DeepSeek} & Astra & Astra & Astra \\
\midrule
Scenes & 13.0 & 0.0 & 1.9 & 0.0 & 1.9 & 0.0 \\
People / organic & 6.0 & 0.0 & 4.0 & 0.0 & 0.0 & 0.0 \\
Objects / 3D & 24.4 & 14.0 & 13.1 & 9.5 & 24.4 & 21.2 \\
Stylized illustration & 35.2 & 7.4 & 11.1 & 18.5 & 24.1 & 11.8 \\
Design / symbolic & 92.0 & 64.0 & 77.3 & 62.0 & 87.3 & 76.5 \\
\midrule
Overall & 47.7 & 28.4 & 34.7 & 28.3 & 42.1 & 39.1 \\
\bottomrule
\end{tabular}}
\caption{Holistic win rates (\%) against GPT-image-1 using the MMRBv2 rubric. Astra denotes GPT-6-Astra; Deepseek denotes Deepseek V4.1 Flash.}
\label{tab:holistic}
\end{table}

%% file: sec/5_harness.tex
\section{Code as an External Visual \texorpdfstring{Language of Thought}{Language of Thought}}
\label{sec:representation}
\label{sec:bottleneck}
\label{sec:harness-effects}
In this section, we first test whether agents can construct and understand visual content through code. Based on the results, we frame code as an \emph{external visual language of thought}: a medium through which agents reason about scenes and apply capabilities such as iterative refinement and in-context learning.

\subsection{Constructing and Understanding through Visual Code}
\label{sec:external-representation}
We test two uses of visual code: constructing an image without rendered-image feedback, and extracting visual information from SVG source code.

\subsubsection{Generation without Visual Feedback}
\label{sec:feedback}

We first ask how well the agent can draw through code and execution feedback alone without visual feedback. We evaluate the GPT-6-Astra agent with Canvas~2D
harness on AgentDrawBench. In the \emph{blind} setting, the
agent receives execution results, but never
sees the RGB canvas. In the \emph{visual feedback} setting, it also
receives the current rendered canvas after each drawing turn.

\paragraph{Results.}
On AgentDrawBench, blind drawing achieves an 83.0\% structural-fidelity pass rate, compared with 85.6\% with visual feedback.
These results support the view that code can serve as a working representation for image construction, with rendered-image feedback providing a modest additional structural gain.

\subsubsection{Understanding Images through Reconstructed Code}

\label{sec:uvg}
We next ask whether agents can interpret visual code to recover information about an image.
To this end, we sample 200 WorldBench image--question pairs proportionally across seven domains~\cite{yin2026worldbench}. The agent reconstructs the original image as SVG before receiving the question. The agent then answers the same question in a fresh context for each
condition, using the original image, the SVG source alone, or the
rendered reconstruction. The code-only condition receives no rendered image.

\begin{figure}[t]
\centering
\includegraphics[width=\columnwidth]{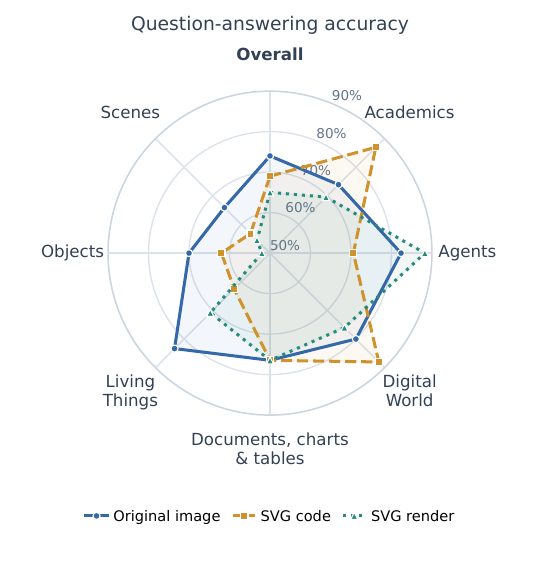}
\caption{WorldBench question-answering accuracy across seven
 domains and overall, using the original image, SVG source code, or rendered SVG reconstruction.}
\label{fig:uvg-domains}
\end{figure}

\begin{figure*}[tbp]
\centering
\includegraphics[width=0.94\textwidth]{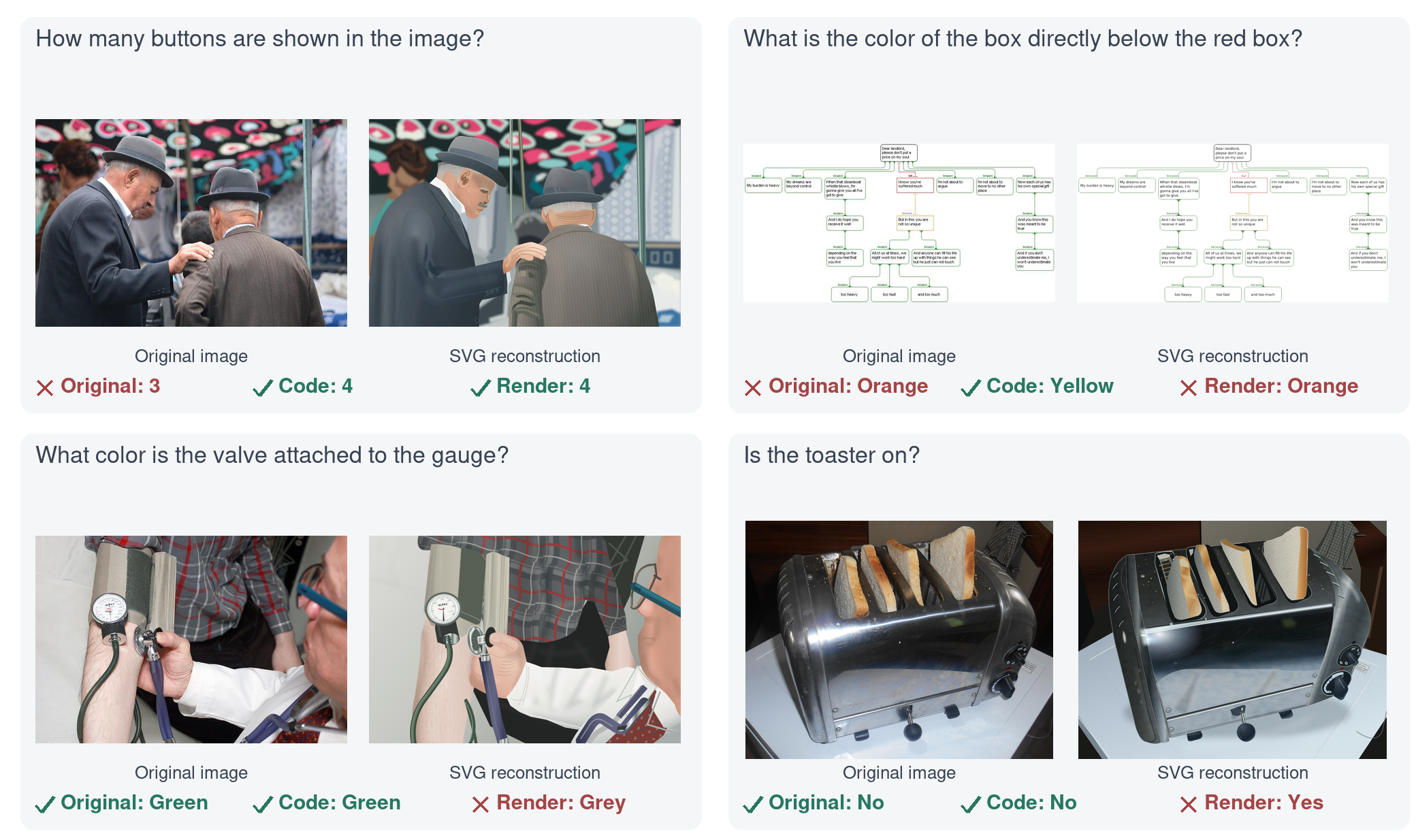}
\caption{Selected WorldBench items where SVG source supports a correct answer while an image-based answer fails.}
\label{fig:uvg-code-successes}
\end{figure*}

As shown in \cref{fig:uvg-domains}, question-answering accuracy is 69.0\% with the reconstruction code and 65.0\% with its render, compared with 74.0\% with the original image. The model answers more questions correctly when given its own program than when given the corresponding rendered image. In the academics and digital-world domains, accuracy from code also exceeds accuracy from the original image. \Cref{fig:uvg-code-successes} illustrates two patterns of successful code-based answering. Code can reduce visual distractions by exposing question-relevant properties separately from surrounding appearance. In the button-counting example, the model correctly identifies four buttons from code despite answering three from the original photograph, as explicit marks are easier to isolate from clothing texture and nearby details. Together, these results support reconstructed code as a usable representation for visual question answering: explicit text, coordinates, and object properties can make question-relevant information easier to access. Details are in \cref{sec:app-uvg}.

\subsection{Agentic Capabilities in Drawing}
\label{sec:external-agentic}
Having shown that agents can construct and understand visual content through code, we next ask whether techniques used in language and code tasks, such as test-time scaling and in-context learning, can also improve drawing with code.

\subsubsection{Test-Time Scaling}
\label{sec:compute}
\paragraph{Experimental Settings}
We test whether additional iterative refinement improves drawing.
We run GPT-6-Astra with Canvas~2D under
standard and extended budgets on AgentDrawBench. For each prompt, the extended run targets a budget of $4B$, where $B=I+5O$ is the measured budget of the standard visual-feedback run, and $I$ and $O$ denote cumulative provider-reported input and output tokens, respectively.  Extended runs allow up to
48 model responses, compared with 12 in the standard setting.
If the model replies \texttt{DONE} before reaching $4B$, the
harness asks it to continue refining the same persistent canvas.

\paragraph{Results.}
On AgentDrawBench, structural fidelity increases from
85.6\% under the standard budget to
86.8\% under the $4\times$ extended budget.
We compare holistic quality directly between the standard and extended runs
using the MMRB2 rubric~\cite{hu2026multimodal}.
Extended refinement achieves a holistic win rate of 50.5\% against standard generation.
Additional refinement within the fixed Canvas~2D harness therefore brings a modest structural gain without a clear holistic preference advantage. This limited return motivates examining the drawing interfaces and procedures available to the agent.

\subsubsection{In-Context Learning for Visual Reconstruction}
\label{sec:style-transfer}
We next examine in-context procedural guidance as a way to extend the drawing interface. Procedures distilled from human demonstrations specify both the marks available to the agent and how they should be organized during reference-guided reconstruction.

\paragraph{Stroke-workflow protocol.}
\label{sec:stroke-protocol}
The agent analyzes sampled frames and stroke sequences from drawing demonstrations, identifying how marks are made and how they accumulate to form shapes. These
observations are translated into reusable stroke operations and
procedural skills specifying construction order, value grouping,
edge control, and local corrections. We develop four workflows:
\emph{Chinese ink} combines tapered lines, broken dry strokes,
dilute washes, and reserved empty space; \emph{charcoal} builds
tonal masses through broad deposits and directional hatching,
then models form through local smudging and lifting;
\emph{watercolor} layers transparent washes, preserves paper
lights, and finishes with selective dry-brush accents; and
\emph{oil painting} progresses from broad color masses to loaded
strokes, glazing, scumbling, and selective impasto. Video sources
and extracted procedures are detailed in
\cref{sec:app-learned-strokes}. For each reconstruction, GPT-6-Astra receives a reference image and the written procedures for the selected medium.  It starts with a blank persistent canvas and refines the drawing using rendered-image feedback.  
The agent may inspect the reference, but its drawing code cannot
sample, trace, blit, or filter the reference pixels.

\begin{figure*}[tbp]
\centering
\includegraphics[width=\textwidth]{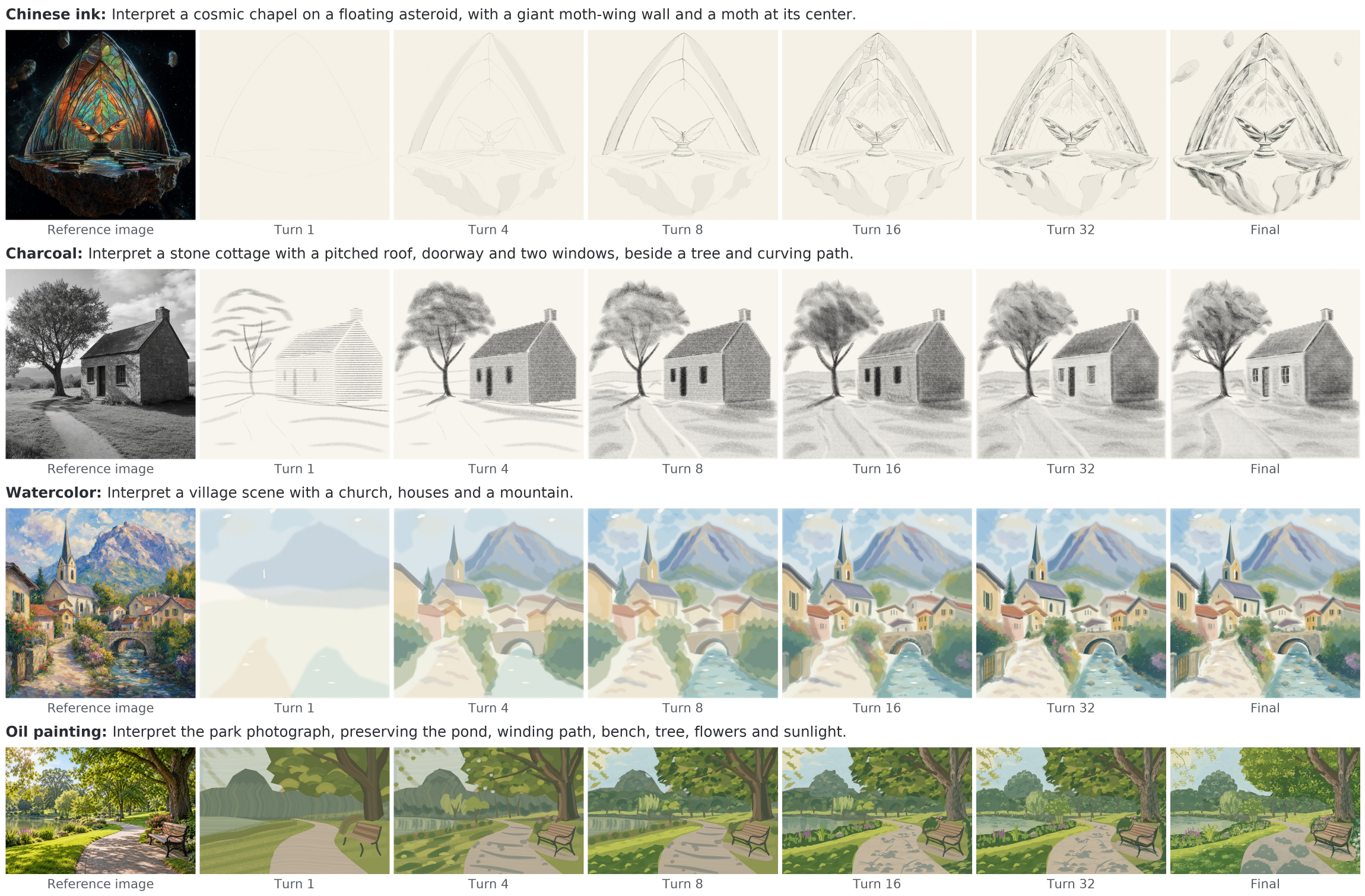}
\caption{Reference-guided image reconstruction using four learned
stroke workflows. Each row shows the reference image,
intermediate renders at turns 1, 4, 8, 16, and 32, and the final
saved image.}
\label{fig:stroke-rollouts}
\end{figure*}

\paragraph{Qualitative reconstruction results.}
The rollouts in \cref{fig:stroke-rollouts} preserve the reference images' recognizable compositions  while producing medium-specific marks: sparse ink
contours in the floating chapel, grainy charcoal tones in the
cottage, transparent watercolor washes in the village, and
layered oil marks in the park. Intermediate frames show construction from broad forms to local detail. These qualitative reconstructions illustrate how procedures supplied in context guide the agent's construction and revision of visual programs to produce expressive, medium-specific effects.

%% file: sec/7_internal_visual_thought.tex
\section{The Internal Visual Language-of-Thought Hypothesis}
\label{sec:inner-vlot}

\begin{figure*}[!tbp]
\centering
\includegraphics[width=\textwidth]{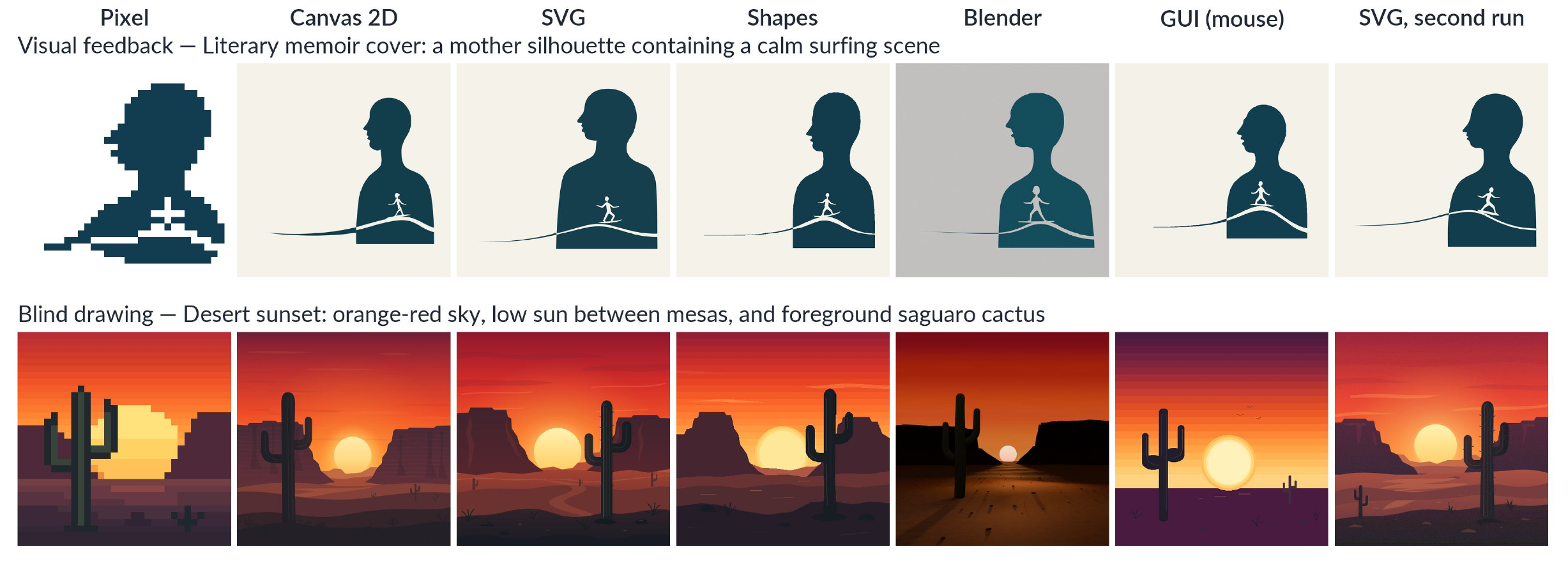}
\caption{Independent GPT-6-Astra drawings for two prompts across six drawing harnesses, including two independent SVG runs. Top: drawings generated with intermediate visual feedback. Bottom: blind drawings of a desert prompt, generated without intermediate visual feedback; blind GUI runs receive only the initial blank window. Pixel provides an additional qualitative comparison. Each run starts from its text prompt without access to another run's code or image. The rows use different prompts and illustrate cross-harness similarities within each setting. Aggregate consistency measurements are reported in \cref{fig:cross-harness-consistency}.  }
\label{fig:cross-harness-200-examples}
\label{fig:harness-ladder}
\end{figure*}

Visual programs allow agents to construct scenes and reason about their contents through an external representation. When comparing their drawings across interfaces, we observe a recurring pattern: independent runs given the same prompt can produce similar overall compositions and visual elements  (\cref{fig:harness-ladder}). Could these recurring visual choices reflect a shared internal representation of scene content? We explore this possibility as the \emph{internal visual language-of-thought hypothesis}.

We hypothesize that agents generate drawings according to the following process: 
\[
I_h = D_h(z),
\]
where $z$ denotes the hypothesized internal representation of scene entities, attributes, and relations, and $D_h$ denotes the process of expressing and rendering that representation through harness $h$ to produce image $I_h$. This effective decoder encompasses the agent's use of drawing programs, GUI actions, or individual pixel responses and their rendering into an image.

Under this hypothesis, drawings produced through different harnesses are different realizations of shared internal imagery. \emph{Consistency} refers to the agreement in visual content across these realizations. \emph{Coherence} is the extent to which underlying order or structure is preserved through change.

We study consistency through independent drawings across harnesses and by independently asking the agent for the color of each pixel on a $128 \times 128$ canvas. We examine the coherence of spatial structure in agent-arranged rooms: the scene arrangement should persist as its appearance changes with viewpoint. We compare room views with floor plans and test the recovery of earlier views after visible drawing programs are removed.

\subsection{Consistency}
\label{sec:consistency}

\begin{figure*}[!tbp]
\centering
\includegraphics[width=\textwidth]{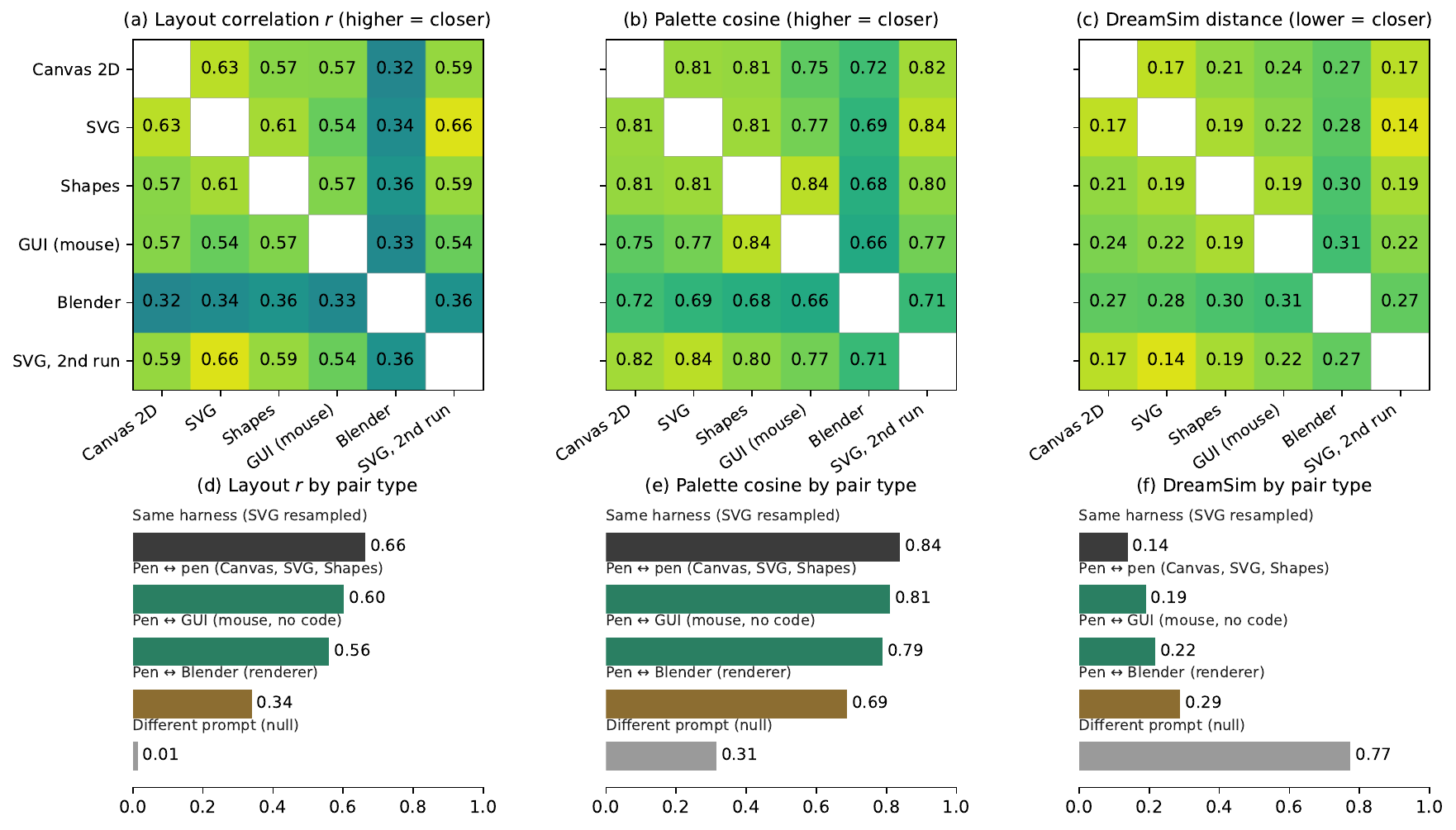}
\caption{Cross-harness consistency on AgentDrawBench. (a--c) Prompt-matched layout correlation, palette cosine similarity, and DreamSim distance between independent drawings of the same prompt using  five harnesses and a second SVG run. (d--f) The same three metrics summarized by pair type. Drawings produced with Canvas 2D, SVG, and Shapes show cross-harness similarity close to that of two independent SVG runs. Mouse-driven GUI drawings show comparable similarity to these outputs. Blender drawings are less similar to the 2D outputs, but remain substantially more similar than drawings of different prompts.}
\label{fig:cross-harness-consistency}
\end{figure*}

\subsubsection{Cross-Harness Consistency}
\label{sec:harness}
We ask whether drawings share visual content when the agent expresses the same prompt through different drawing interfaces. We compare independent drawings on AgentDrawBench through Canvas~2D, SVG, Shapes, the mouse-driven GUI, and Blender, with a second independent SVG run as a same-harness reference. Each run starts from the prompt without access to another harness's program or image. We measure coarse luminance layout, dominant hues, and perceptual similarity. A separate ten-prompt pilot tests cross-harness agreement without visual feedback; \cref{fig:harness-ladder} illustrates selected outputs.

\paragraph{Luminance layout fingerprints.}
We convert each output to grayscale using
$L=0.299R+0.587G+0.114B$, downsample it to $16\times16$ with
Lanczos filtering, and flatten the result into a 256-dimensional
vector. This coarse representation suppresses fine textures and
stroke details while retaining the spatial arrangement of large
light and dark regions. For two drawings of the same prompt,
we compute the Pearson correlation between their vectors:
\begin{equation}
 r(x,y)=\frac{\sum_i(x_i-\bar{x})(y_i-\bar{y})}
 {\sqrt{\sum_i(x_i-\bar{x})^2\sum_i(y_i-\bar{y})^2}}.
\end{equation}
This measures the agreement in coarse tonal layout, rather
than the identity of the object or exact geometric agreement.

\paragraph{Saturation-weighted hue palettes.}
We convert each full-resolution RGB image to HSV and
construct an 18-bin hue histogram. Each pixel contributes its saturation to the corresponding hue bin, giving greater weight
to saturated colors than to nearly gray regions. For hue $H_p$ and saturation $S_p$ at pixel $p$,
the histogram is
$h_b=\sum_p S_p\,\mathbf{1}[H_p\in B_b]$,
where $B_b$ is one of 18 equal hue intervals.
We compare histograms using cosine similarity,
$c(h,g)=h^\top g/(\|h\|_2\|g\|_2)$.
This representation discards spatial position and measures color usage.

\paragraph{DreamSim~\cite{fu2023dreamsim}}
We also report the DreamSim distance between independent drawings of the same prompt. DreamSim is a learned perceptual metric trained to align with human similarity judgments, capturing differences in semantic content, object appearance, color, and spatial layout. Lower values indicate greater perceptual similarity.

\paragraph{Results and controls.}
\revadd{\Cref{fig:cross-harness-consistency} summarizes the prompt-matched similarities on AgentDrawBench. Three comparisons highlight the main findings. First, drawings of the same prompt using two different 2D drawing harnesses (Canvas~2D, SVG, Shapes) agree at a mean layout correlation of $0.60$, palette cosine $0.81$, and DreamSim $0.19$, against $0.66$, $0.84$, and $0.14$ for two independent SVG runs. Thus,  cross-harness agreement remains close to the same-harness reference on all three metrics. Second, the computer-use harness, which shares no drawing API with the others and draws by mouse, sits inside this cluster ($0.56$, $0.79$, $0.22$). Third, Blender, which constructs and renders a 3D scene, is further from every 2D drawing harness ($0.34$, $0.69$, $0.29$), while still showing substantially greater similarity than the different-prompt null ($0.01$, $0.31$, $0.77$). The same ordering holds across the three metrics. A ten-prompt pilot without visual feedback gave the same ordering (layout correlations of  $0.42$--$0.49$ compared with   a null of $0.03$--$0.08$). These independent runs show that coarse layout and appearance choices recur across different ways of constructing an image, including mouse-driven drawing. Their agreement supports the hypothesis that drawings produced through different interfaces reflect shared internal imagery}.

\subsubsection{Independent Queries, Consistent Images}
\label{sec:pixel-probe}
\begin{figure*}[!tbp]
\centering
\includegraphics[width=\textwidth]{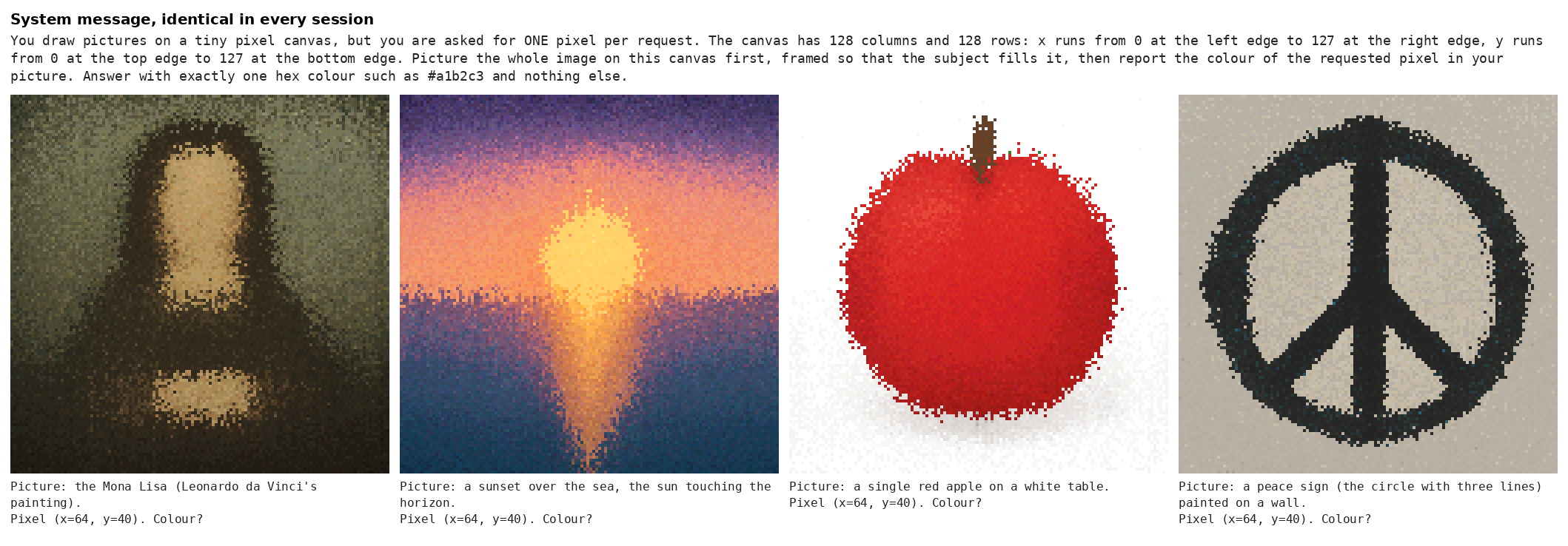}
\caption{\revadd{Pictures assembled from independent sessions: the Mona Lisa, a sunset over the sea, a red apple, and a peace sign, each pixel answered in a separate context with no access to any other answer. The system message shown is identical in every session; the user message names the subject and one coordinate. Layout and palette recur across sessions; detail and exact pixel values do not.}}
\label{fig:pixel-probe}
\end{figure*}

\revadd{Whole-image generation allows the agent to coordinate scene elements within a drawing program. We next ask whether consistent visual structure also emerges when the agent is asked for the color of each pixel independently. We ask the agent to ``draw'' a subject on a $128{\times}128$ canvas but to report only the color of one pixel $(x,y)$. Every pixel is requested in a separate, fresh context, without access to any other query's answer; the 16{,}384 answers are then assembled. \Cref{fig:pixel-probe} shows four images assembled this way.}

The independently queried pixels assemble into the coarse composition of the Mona Lisa: a light oval face framed by dark hair, a dark torso, and a lighter patch in the hand region (\cref{fig:pixel-probe}). These regions occupy compatible positions within a recognizable whole, even though each query has no access to any other answer. Facial features remain unresolved and boundaries are noisy; the agreement lies in the broad spatial organization of the portrait.

\revadd{The agreement occurs at the level of regions and tones across answers generated without a shared drawing program or access to one another's outputs. The consistency is therefore among independently generated parts of the assembled image, supporting the hypothesis that local responses can express shared internal imagery.}

\subsection{Coherence}
\label{sec:coherence}
Consistency across expressions motivates a further test: can the agent preserve a scene's spatial structure through successive viewpoint changes and revisits? We examine spatial coherence in self-generated rooms. Given a predefined furniture list, the agent chooses where to place the furniture, then draws the room from different headings and returns to earlier views. The room's arrangement should remain stable while visibility and image-relative ordering change with viewpoint, and revisits should recover the earlier views. By removing earlier SVGs from the visible conversation, we test whether the agent can continue to express its own scene choices without rereading those drawings.

\begin{figure*}[!tbp]
\centering
\includegraphics[width=\textwidth]{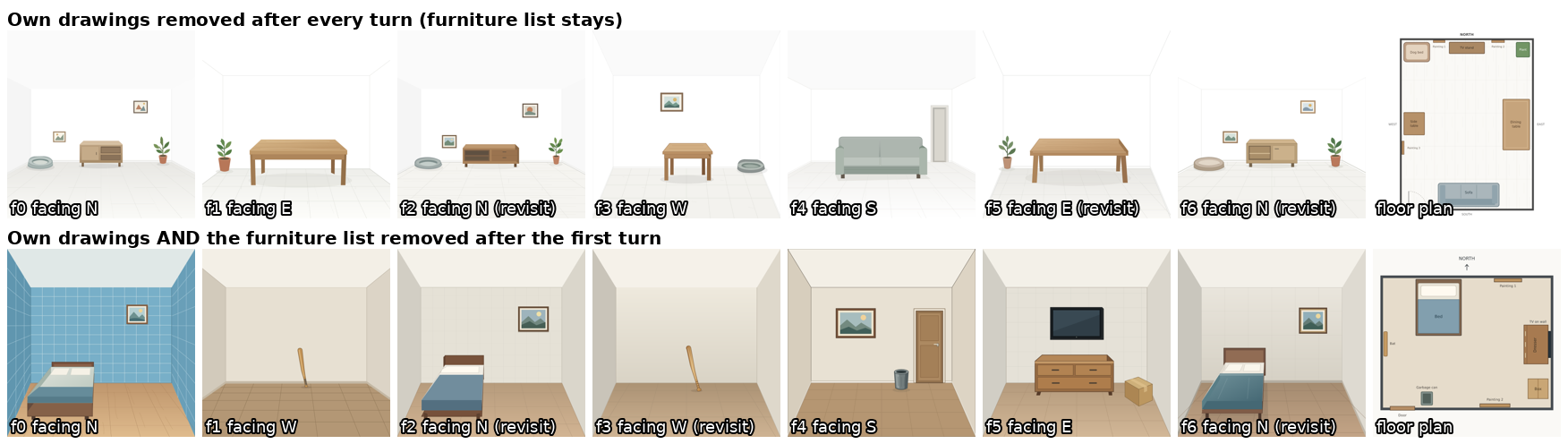}
\caption{\revadd{Self-furnished rooms, two conversations. The agent is given only a furniture list, arranges the room itself, draws the facing wall, turns six times, and ends with a floor plan; the frames are shown in the order drawn, revisited headings marked. Top: its own drawings are deleted from the context after every turn. Bottom: the furniture list is deleted too; the same pieces return on the same walls and the plan lists all of them.}}
\label{fig:self-rooms}
\end{figure*}

\paragraph{Test setting.}
\revadd{The agent receives the type and rough shape of one of 30 rooms from the ProcTHOR-10K validation split~\cite{deitke2022procthor} and a list of its 7--14 furniture pieces with size words but no positions. It is told to decide an arrangement and keep it fixed, to stand at the centre, and to draw what it sees as SVG with one group per piece. It draws the facing wall, then follows a random sequence of six quarter- and half-turns that visits every wall, revisits headings, and ends facing north, and finally draws a top-down floor plan (\cref{fig:self-rooms}).}

\revadd{Four conditions vary the retained context: (1) the agent's own drawings (SVG) are kept; (2) they are replaced by one-line stubs; (3) both drawings and the furniture list are replaced by stubs; or (4) every view is a fresh call given only the list and the turn history. Every condition covers all 30 rooms. }

\revadd{With no externally specified target arrangement, we score agreement among the agent's own outputs geometrically from the SVG: whether a revisited heading shows the same pieces and relations as its first view (piece-set Jaccard, relation agreement), and whether each wall view shows exactly the pieces the agent's own plan places on that wall, in the same left-to-right order. Chance is obtained by permuting the plan's wall assignment. All conditions use GPT-6-Astra with SVG output; protocol details are in \cref{sec:app-selfrooms}.}

\paragraph{Results.}
\revadd{\Cref{tab:self-rooms} gives the results across four conditions. With its earlier drawings deleted after every turn, the agent still redraws the same room on each revisit (Jaccard $0.99$, relation agreement $0.96$) and its wall views agree with its own plan on 99\% of pieces (chance $0.31$); box IoU is $0.71$, against exactly $1.00$ when the drawings are kept. When the furniture list is deleted as well, removing that explicit record of the room contents, the agent retains most furniture pieces and their wall assignments (Jaccard $0.88$, plan agreement $0.99$), including every one of the 13 pieces in the illustrated room (\cref{fig:self-rooms}, bottom).}

\begin{table}[t]
\centering
\small
\setlength{\tabcolsep}{3pt}
\begin{tabular}{@{}p{0.42\columnwidth}cccc@{}}
\toprule
Context after the first view & \shortstack{Same\\pieces} & Relations & \shortstack{Plan\\wall} & Order \\
\midrule
Own drawings kept & 0.97 & 1.00 & 0.99 & 0.92 \\
Own drawings removed & 0.99 & 0.96 & 0.99 & 0.79 \\
Drawings and list removed & 0.88 & 0.79 & 0.99 & 0.80 \\
Fresh call per view & 0.71 & 0.61 & 0.24 & 0.63 \\
\midrule
Chance & -- & -- & 0.31 & 0.50 \\
\bottomrule
\end{tabular}
\caption{\revadd{Self-furnished rooms: revisit consistency (piece-set Jaccard and relation agreement between a revisited heading and its first view) and agreement between each wall view and the agent's own floor plan (piece on exactly its plan wall; left-to-right order). 30 rooms.}}
\label{tab:self-rooms}
\end{table}

\paragraph{Interpretation.}
\revadd{The room experiment shows that later drawings remain compatible with the agent's earlier scene choices as the requested viewpoint changes. Furniture positions were never supplied in the prompt, and earlier SVGs are unavailable in the removal conditions, so compatible revisits and floor plans suggest  the continued availability of the agent's chosen arrangement.  This preservation of scene structure is the spatial coherence expected under the internal visual language-of-thought hypothesis. Across the three probes, visual structure recurs under changes of interface, independent queries for pixel colors, and changes of viewpoint. Together, these behaviors support the hypothesis of reusable scene content underlying external visual expressions.}

%% file: sec/7_discussion.tex
\section{Conclusion}
\label{sec:discussion}
General-purpose coding agents can draw zero-shot, with particular strengths in structural accuracy and compositional constraints and persistent limitations in perceptual richness. Drawing harnesses shape how these capabilities appear in the resulting images. Visual programs also serve as an external visual language of thought: agents can construct them and interpret the scene information they express. Beyond these observable operations, cross-harness consistency, independent pixel queries, and coherence in self-furnished rooms provide complementary behavioral evidence for an internal visual language-of-thought hypothesis. Together, these results connect agents' use of executable visual representations to a possible shared organization of scene content across external expressions.

\paragraph{Limitations.}
Most ablation experiments use only one primary agent, and evaluations largely rely on automated judgments with limited human verification. The in-context learning experiments are evaluated qualitatively. We study frontier agents through the interfaces available to us, without access to model weights or the providers' internal agent interfaces and implementations. Our investigation of the internal visual language-of-thought hypothesis is therefore exploratory and behavioral: consistency and coherence provide supporting evidence. The pixel-query probe covers five subjects, and the self-furnished-room experiment concerns an agent with retained, opaque reasoning history. The experiments cover selected drawing harnesses and spatial transformations; broader evaluation across models, interfaces, and transformation types would provide a more comprehensive account of these capabilities.

%% file: sec/appendix.tex
\clearpage
\ifdefined\arxivversion
\section*{Appendix}
\else
\renewcommand{\thepage}{S\arabic{page}}
\setcounter{page}{1}
\maketitlesupplementary
\fi
\appendix



\section{Harness and Run Infrastructure}
\label{sec:app-harness}

\subsection{The harnesses}

\begin{table}[h]
\centering
\small
\resizebox{\columnwidth}{!}{%
\begin{tabular}{llp{3.4cm}}
\toprule
Harness & Renderer & Representation / interaction \\
\midrule
Pixel & PIL & Individual pixel assignments \\
Shapes & PIL & Ordered geometric marks \\
Canvas~2D & Node + Canvas~2D & JavaScript on a persistent canvas \\
SVG & resvg & Complete vector-document replacement \\
GUI & JS Paint / Chromium & Mouse and keyboard actions with screenshots \\
Blender Cycles & Cycles & Procedural 3D scenes with render feedback \\
Stroke workflows & Canvas~2D & Medium-specific marks and construction procedures \\
SVG patch & resvg & Incremental append and replace-by-id control \\
\bottomrule
\end{tabular}%
}
\caption{Drawing harnesses, renderers, and interaction mechanisms. Their protocols
are not uniformly matched in compute or feedback.}
\label{tab:app-harness}
\end{table}

\paragraph{pixel.} A single tool,
\texttt{draw\_pixel(drawing\_index, color)}, over a PIL RGB
canvas. The address is a flat row-major index ($0=$ top-left);
colors are CSS names or \texttt{\#RRGGBB}; out-of-range indices
and unparseable colors come back as tool errors.

\paragraph{Shapes.}
The model draws a sequence of elementary geometric marks:
filled rectangles, ellipses, polygons, lines with adjustable
width, and individual pixels. It specifies each mark's geometry
and color; coordinates are rounded to integer pixel positions.
Marks accumulate on the same raster canvas in the order generated,
so later marks can cover earlier ones.

\paragraph{Canvas~2D.}
The model writes executable JavaScript programs that draw onto
a persistent canvas. Programs can combine paths, curves, text,
gradients, transformations, and compositing, while loops and
reusable routines express repeated elements and procedural
textures.

\paragraph{SVG and incremental SVG editing.}
The model constructs a vector document describing the image's
shapes, paths, text, colors, gradients, transformations, and
layering. The standard SVG harness replaces the complete document
at each revision. An incremental control variant instead allows
new elements to be appended and existing elements to be replaced
by identifier.

\paragraph{GUI.}
\label{sec:app-gui}
The GUI harness runs JS Paint in headless Chromium through
Playwright. The model sends ordered batches of clicks,
double-clicks, mouse-drag paths, typed text, key presses, and color
selections in window coordinates. A color-selection shortcut
opens and operates the application's Edit Colors dialog.
Available application tools include pencil, brush, airbrush,
eraser, fill, lines, curves, rectangles, ellipses, polygons, and
text. Tool choice, shape options, and canvas dimensions are set
through the interface; undo is available through keyboard input.
After each batch the agent might receive a screenshot containing both
the canvas and application controls. 
Code execution, image import, and paste are unavailable
as agent actions.

\paragraph{Blender Cycles.}
\label{sec:app-blender-settings}
CPU rendering uses adaptive sampling, OIDN denoising, and the AgX
view transform. Preview and final timeouts are 60 and 180 seconds,
respectively, within a cumulative rendering budget of 360 seconds.
Each script rebuilds the scene; external assets are excluded.
Of 200 AgentDrawBench prompts, 199 have successful final renders;
one exhausts the render-time budget. Structural evaluation reuses
cached judgments of these exact images. Perceptual and holistic
comparisons against GPT-image-1 use the same applicable dimensions
and MMRBv2 rubric as the other harnesses, with Gemini~3.5~Flash
through the Arena gateway and both image orderings. Perceptual
judging requests all applicable dimensions in one JSON response
per ordering. Different response limits and rendering costs make
this a comparison of harness configurations rather than a
controlled estimate of the renderer's causal effect.

\paragraph{Stroke workflows.}
\label{sec:app-learned-strokes}
\revadd{We inspect sampled video frames and short stroke sequences
and encode the observed ordering, mark families, and correction
procedures as instructions and medium-specific Canvas~2D helpers.
This is in-context procedural guidance. Reference-guided studies
begin on a blank canvas; the model may inspect the supplied
reference and its current render, but its drawing code cannot
sample, trace, blit, or filter reference pixels. Intermediate
renders and executed programs are retained. These qualitative
studies do not establish a controlled improvement on
AgentDrawBench.}

\subparagraph{Oil.}
\revadd{The source is the oil-painting timelapse shared by blueemi99,
\url{https://x.com/blueemi99/status/2106755903223664708}.
We analyze a recording of this demonstration.
The visible timelapse proceeds from a toned ground and composition
scaffold to broad light/color masses, structural forms,
material-directed strokes, and sparse focal accents. We adapt
this ordering with loaded and dry strokes, glazing, scumbling,
local blending, and selective impasto.}

\subparagraph{Watercolor.}
\revadd{Sources are Javid Tabatabaei's landscape demonstration,
\url{https://www.youtube.com/watch?v=wnG6vg3TRlE};
misulbu's portrait demonstration, \emph{watercolor painting--When overlapping},
\url{https://www.youtube.com/watch?v=jCJsnpH9IC4}; and
Tabatabaei's \emph{Watercolor painting tutorial--a Rainy Night},
\url{https://www.youtube.com/watch?v=GjVGfTPqaoA}.
The workflow plans light reserves and broad color relationships,
then layers transparent masses, strengthens connected shadows,
and adds selective dry-brush and focal accents. Portrait studies
prioritize face/hand proportions before fine details; rainy-street
studies align broken pavement reflections with their light sources.
Rendering controls approximate wetness, grain, and edge softness.}

\subparagraph{Charcoal.}
\revadd{The source is Kisoo Chai's \emph{Real-time Charcoal Drawing, \#157},
\url{https://www.youtube.com/watch?v=K-Ane8Bq7gY}.
The observed sequence establishes placement and feature axes,
groups dark masses, models turning planes, then alternates local
smudging, lifting, and selective accents. Our implementation uses
broad granular deposits, directional hatch bundles, local pigment
redistribution, and partial removal. Fine features follow connected
midtone coverage; repeated contour reinforcement is replaced by
lifting and rebuilding the faulty plane.}

\subparagraph{Chinese ink.}
The source is the Chinese ink demonstration on Bilibili,
\url{https://www.bilibili.com/video/BV1sg411Q7av/}.
We analyze \revblue{a recording} of this demonstration.
The sampled demonstration establishes pavilion contours, branching
pines and grouped needles, broken foreground/shore marks,
pale mountain washes, and selective dark ridge accents while
leaving substantial sky and water as untouched paper. We adapt
its tapered pressure lines, form-following dry texture, dilute
washes, and dense-versus-empty balance. Chinese ink specifies the medium, while the reference image determines the subject matter.

\subsection{System-prompt skeleton and turn mechanics}
\label{sec:app-skeleton}
\paragraph{Generator, turns, and stopping.}
A turn is one API call; the model may attach any number of tool
calls, which the harness executes in listed order, returning all
results in one message. A reply of \texttt{DONE} with no tool
calls ends the run. A stop without
\texttt{DONE} is answered with a nudge to continue or finish
explicitly, twice by default, before the run is recorded as
\texttt{stopped}. Turn caps are 12 for the code harnesses and 40
for \texttt{shapes}.

The core programmable harnesses receive instructions describing
the output device, visibility, canvas dimensions, available drawing
operations, construction strategy, and stopping rule. Dimensions,
background color, and the visibility clause are instantiated for
each run. Below we reproduce the construction and stopping
instructions from the system prompts; the operation declarations
are described in the preceding harness specifications.

\paragraph{Pixel system-prompt excerpt.}
\begin{quote}\small
Plan the picture first (briefly), then paint it. Fill in every pixel
that should differ from the background. When the picture is complete,
reply with the single word DONE and no tool calls.
\end{quote}

\paragraph{Shapes system-prompt excerpts.}
\begin{quote}\small
Later calls paint over earlier ones, so work back to front: sky and
ground first, then distant objects, then foreground detail.
Tool calls within one turn are executed strictly in the order you
list them, so issue many calls per turn (dozens is fine); one call
per turn is far too slow.

Plan the composition briefly, then paint it. When the picture is
complete, reply with the single word DONE and no tool calls.
\end{quote}

\paragraph{Canvas~2D system-prompt excerpt.}
\begin{quote}\small
Work back to front in a few substantial scripts: background and sky,
terrain and large masses, mid-ground, foreground detail, atmosphere.
Use gradients, paths, transforms and loops to get real richness at
this resolution; avoid a few flat rectangles. Plan briefly first.
When the picture is complete, reply with the single word DONE and
no tool calls.
\end{quote}

\paragraph{SVG system-prompt excerpt.}
\begin{quote}\small
Plan briefly, then write a rich, well-layered SVG (background,
large masses, mid-ground, detail). When the picture is complete,
reply with the single word DONE and no tool calls.
\end{quote}

Small wording choices in this skeleton carry measurable weight.
An early shapes prompt lacked the sentence promising that tool
calls within one turn are executed strictly in the order listed;
without it, the generator issued one primitive per turn, and 3 of 9 pilot runs hit the turn cap
with unfinished images. With the sentence, the same model issued
20--64 calls per turn and no run hit the cap. The added sentence makes the harness's execution order explicit.

\paragraph{A representative rollout.} AgentDrawBench prompt
\texttt{arena\_132} (a social-media post mock-up with exact
headline text) under arm~C: turn~1 plans in prose, then lays the
background and card geometry in one script; turn~2, before
committing the headline, \emph{measures} it ---
the rendered width of the headline over two
candidate font stacks (583.8 vs.\ 581.2\,px against the intended
column) --- and only then draws; turns~3--5 add body text,
avatar, and interaction row, each checking the returned render;
the run ends with \texttt{DONE} after 5 turns at \$0.58. This
measure-then-commit pattern recurs throughout the canvas2d traces and illustrates how executable code supports measurement during image construction.

\subsection{Imposed-budget mode}
\label{sec:app-budget}

The dollar-matched arms of \cref{sec:compute} use
\texttt{--min-spend}/\texttt{--max-spend}. Below the floor, a
\texttt{DONE} is not accepted: the harness replies with a nudge of
the form ``You still have about $X\%$ of your drawing budget
left. Keep refining the same canvas\ldots'' and the run
continues. The ceiling is checked before each turn; crossing it
stops the run . Spend is the
gateway-reported dollar cost accumulated so far, so matched arms
are matched in dollars, not in tokens or turns. The
\texttt{--feedback text} mode (a character grid downsampled from
the render) is image-derived, and is therefore excluded from
every no-inspection arm.

\subsection{Run records, resumability, and cost accounting}
Each run preserves the final image, vector source where applicable,
intermediate frames, a timelapse, the complete interaction trace,
the instantiated system prompt, the task, and judge outputs.
Records include response counts, stopping reason, elapsed time,
provider-reported token usage and cost, and drawing-operation counts.
Batch generation and judging skip completed records when resumed.
Costs are accumulated from provider-reported usage.

\section{WorldBench Selection and Paired Answer Analysis}
\label{sec:app-uvg}

The frozen set contains 200 image--question items from the
\texttt{zlab-princeton/WorldBench} train split. The documented sampling procedure is random and stratified proportionally by
domain with seed 20261001. The selection script itself is not
tracked, so the frozen manifest \revblue{defines} the exact selection. Its entries retain the
source row index, domain, category, question, four answer options,
and benchmark answer key. This analysis uses the frozen sample and recorded model responses. Domain allocations are Objects 50, Scenes 44,
Digital World 25, Living Things 24, Academics 23, Agents 17, and
Documents/Charts/Tables 17. The following source row indices identify
every selected item, grouped by domain.

\paragraph{Academics} 1678, 1688, 1700, 1703, 1707, 1717, 1718, 1730, 1752, 1762, 1770, 1788, 1805, 1819, 1824, 1832, 1845, 1855, 1859, 1866, 1875, 1882, 1888.

\paragraph{Agents} 1342, 1385, 1388, 1390, 1395, 1399, 1401, 1403, 1404, 1416, 1423, 1433, 1436, 1451, 1458, 1476, 1480.

\paragraph{Digital World} 14, 27, 31, 46, 48, 93, 98, 100, 109, 116, 125, 127, 129, 1903, 1913, 1916, 1925, 1930, 1938, 1961, 1964, 1980, 1993, 1995, 1997.

\paragraph{Documents, Charts, \& Tables} 1506, 1515, 1536, 1559, 1563, 1565, 1571, 1575, 1615, 1616, 1618, 1624, 1627, 1637, 1647, 1651, 1900.

\paragraph{Living Things} 1099, 1106, 1109, 1116, 1124, 1137, 1139, 1140, 1141, 1170, 1173, 1183, 1201, 1203, 1208, 1213, 1223, 1235, 1257, 1258, 1286, 1288, 1302, 1331.

\paragraph{Objects} 608, 612, 619, 651, 658, 677, 679, 682, 684, 696, 699, 700, 703, 719, 742, 764, 769, 774, 776, 811, 813, 823, 824, 839, 845, 848, 858, 862, 863, 864, 866, 890, 908, 921, 929, 933, 937, 950, 973, 976, 990, 994, 1000, 1009, 1021, 1040, 1067, 1068, 1073, 1243.

\paragraph{Scenes} 159, 169, 173, 202, 222, 227, 228, 230, 233, 242, 244, 256, 261, 263, 264, 279, 293, 316, 356, 357, 384, 393, 396, 424, 427, 428, 434, 437, 448, 458, 459, 467, 474, 475, 490, 499, 505, 510, 537, 544, 555, 567, 574, 575.

\paragraph{Conditions and analysis.}
The reconstruction is generated from the source image before the
quiz; the reconstruction request does not contain the MCQ.
Separate fresh-context calls answer the same benchmark question
from the original image, rendered reconstruction, or SVG source.
The question-answering evaluation uses \texttt{gpt-6-astra}; this is a
single-model analysis. SVG input is truncated at 60,000 characters
by the runner. The matching
experiment has 145 items, as collection stopped at the budget limit;
it supplies the SVG plus its own render or a randomly selected
other item's render. It is separate from the 200-item MCQ analysis.

We compare exact-match correctness against the benchmark answer
key. For domain $d$, the original-to-render loss rate is
\[
\ell_d =
\frac{\sum_{i\in d}\mathbf{1}[O_i=1,R_i=0]}
     {\sum_{i\in d}\mathbf{1}[O_i=1]},
\]
where $O_i$ and $R_i$ denote correctness from the original and
render. This conditional rate separates reconstruction-associated
answer changes from errors already present with the original.
There are 32 losses and 14 reverse gains, yielding a net decrease
of 18 correct answers (9 percentage points). The analogous
original-to-code comparison has 30 losses and 20 gains; code-to-render
has 19 losses and 11 gains. These are paired response transitions,
not proof that the underlying visual fact is absent from an artifact.
MCQ guessing, visual reasoning, and code reasoning can all affect
correctness. Domain sample sizes are small; the displayed ranking
is descriptive, with no claim of statistical significance.

\begin{figure}[t]
\centering
\includegraphics[width=\linewidth]{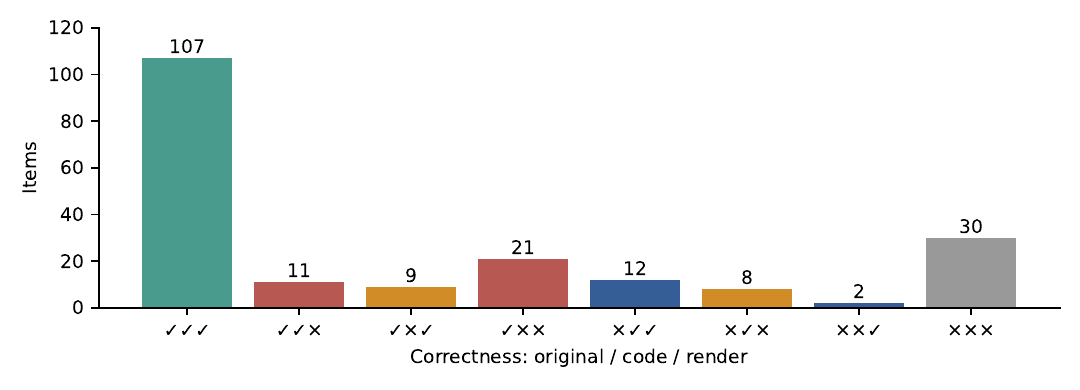}
\caption{Complete correctness census for the 200 WorldBench items.
Each label gives original/code/render correctness in that order.
There are 107 all-correct, 30 all-wrong, and 63 mixed items.}
\label{fig:uvg-transitions}
\end{figure}

The paired analysis produces the correctness-census plot and a complete per-item record of answer correctness.

\section{Independent Pixel Queries: Additional Diagnostics}
\label{sec:app-pixel-diagnostics}
The following diagnostics supplement the qualitative analysis in \cref{sec:pixel-probe}. Across five subjects, the assembled pictures have high spatial autocorrelation (Moran's $I$ of luminance $0.85$--$0.96$, against $0.01$ when the same colors are shuffled) and share coarse composition with whole-image Canvas~2D drawings of the same subjects (luminance correlation $0.51$--$0.88$). Re-asking the same coordinate returns the identical color only 2--38\% of the time, though the difference is small (about 10 of 255 per channel), and fine structure such as facial features or the leaves of a branch does not survive.

\section{Self-Furnished Room Protocol}
\label{sec:app-coherence}
\subsection{Self-Furnished Rooms}
\label{sec:app-selfrooms}
The experiment in \cref{sec:coherence} uses 30 rooms
from the ProcTHOR-10K validation split. The agent receives a room
type, rough shape, and a list of 7--14 furniture pieces described
with size words, without furniture positions. It chooses an
arrangement, draws an initial wall view as SVG with one group per
piece, and follows six quarter- and half-turns. The turn sequence
visits every wall, revisits headings, and ends facing north. A
final top-down plan provides a second expression of the agent's
chosen arrangement.

The four conditions retain earlier SVGs, replace them with
one-line stubs, replace both the SVGs and the furniture list with
stubs, or use a fresh
call for each view with only the furniture list and turn history.
Every condition covers all 30 rooms.

Scoring measures self-consistency because the agent chooses the
layout. Piece-set Jaccard and geometric relation agreement compare
a revisited heading with its first view. Object-box IoU measures
the stability of placement on revisits. Wall--plan agreement
checks whether a piece appears on the wall assigned by the agent's
own plan; order agreement compares left-to-right order within a
wall. Permuting the plan's wall assignments provides the chance
reference. These comparisons assess persistence and coordination
of the agent's chosen scene, rather than accuracy against an
externally supplied arrangement.